\documentclass[10pt]{article}
\usepackage[preprint]{tmlr}

\usepackage{amsmath,amsfonts,bm}

\def\eqref#1{equation~\ref{#1}}

\def\1{\bm{1}}

\DeclareMathAlphabet{\mathsfit}{\encodingdefault}{\sfdefault}{m}{sl}
\SetMathAlphabet{\mathsfit}{bold}{\encodingdefault}{\sfdefault}{bx}{n}

\usepackage{amsmath}
\usepackage{booktabs}
\usepackage{graphicx}
\usepackage{hyperref}
\usepackage{microtype}
\usepackage{multirow}
\usepackage{placeins}
\usepackage{url}
\usepackage{xcolor}
\hypersetup{colorlinks=true,linkcolor=blue,citecolor=blue,urlcolor=blue}

\title{A Controlled Evaluation of Curvature-Inspired\\
Scalar Feature Scaling for Domain Generalization}

\author{%
\name Sunny Gupta \email sunnygupta@iitb.ac.in\\
\addr Indian Institute of Technology Bombay\\
Mumbai, Maharashtra, India
\AND
\name Amit Sethi \email asethi@iitb.ac.in\\
\addr Indian Institute of Technology Bombay\\
Mumbai, Maharashtra, India
}

\def\month{August}
\def\year{2026}
\def\openreview{Under review}

\newcommand{\method}{\textsc{HyperDG}}
\newcommand{\pp}{\,pp}
\newcommand{\full}{\textsc{FullPerDomain}}
\newcommand{\erm}{\textsc{StandardERM}}
\newcommand{\shared}{\textsc{SharedLearnable}}
\newcommand{\novar}{\textsc{NoVariance}}

\begin{document}
\maketitle
\footnotetext{Generative-AI tools were used for language editing and
typesetting assistance. The authors verified the analysis, claims, citations,
and final text and take responsibility for the manuscript.}

\begin{abstract}
Domain-specific curvature is an appealing idea for domain generalization (DG),
but \emph{curvature learning} can denote substantially different computations.
This paper evaluates one deliberately narrow implementation: the code calls it
\method{}, but it rescales Euclidean encoder features during training by
learned positive scalars and predicts with an unscaled Euclidean feature and
an ordinary linear classifier. It does not implement a Lorentz or
Poincar\'e representation, geodesic prediction, or a Sharpness-Aware
Minimization inner loop. We audit the executed objective and complete a
six-condition, four-dataset (PACS, VLCS, OfficeHome, TerraIncognita),
four-target, four-seed evaluation (384 runs), with checkpoints selected using
only the three source-validation environments and target-level effects, not
16 seed-by-target cells, as the per-dataset inferential unit. The full
implementation's equal-dataset macro estimate is $1.61$ percentage points
below native DomainBed ERM; all four dataset estimates favor ERM, but the
primary four-dataset exact sign-flip test gives $p=.125$ (Holm-adjusted
$p=.250$), so we claim no adjusted evidence of a difference. The cleaner
domain-specificity contrast, separate learned source scales without their
dispersion penalty versus one shared learned source scale, is $-0.13\pp$ and
passes a revision-stage $\pm1\pp$ equivalence test (TOST $p=.003$) for the
source-selected suite macro, though not for every dataset, target, or the
final-step endpoint. The study thus supplies a bounded negative result for
this scalar training package; it does not test or refute genuine hyperbolic
representation learning.
\end{abstract}

\section{Introduction}
\label{sec:introduction}

Domain generalization (DG) trains on several source environments and evaluates
on an unseen target environment \citep{wang2022generalizing,gulrajani2021insearch}.
The setting invites geometric inductive biases: hyperbolic spaces can represent
hierarchies and multiple scales compactly, and modern hyperbolic networks can
operate directly on a curved manifold \citep{ganea2018hyperbolic,
fan2025curvaturelearning}. A much weaker intervention is to multiply an
otherwise Euclidean feature vector by a learned scalar. Those interventions
should not be evaluated or described as if they were interchangeable.

This distinction is central here. \emph{In this paper,} \method{} denotes the
exact scalar-scaling formulation specified in Section~\ref{sec:method}; it is
not a reference to a separate external paper and it is not shorthand for
hyperbolic DG in general. The implementation maps a Euclidean feature
$h_\theta(x)$ to
$h_\theta(x)/\sqrt{c_i}$ during its training loss, where $c_i>0$ is indexed by
a source-loader slot. The model predicts with the unscaled Euclidean feature
and a linear classifier. Calling $c_i$ a curvature magnitude reflects the
implementation's motivation, not a claim that the representation has learned
sectional curvature.

We ask two bounded questions:
\begin{enumerate}
    \item Does the complete \method{} training package improve held-out-domain
    accuracy over native DomainBed ERM under a matched ResNet-50 protocol?
    \item Within the same \method{} update route, do three separately learned
    source scales provide a practically meaningful macro advantage over one
    shared learned source scale?
\end{enumerate}
The first comparison is a whole-package utility test, because native ERM and
\method{} differ in update and BatchNorm routes. The second isolates the
per-source parameterization more cleanly by comparing \novar{} with \shared{};
neither condition has an operative between-scale dispersion penalty.

This revision contributes:
\begin{itemize}
    \item an implementation-level specification that separates the intended
    geometric interpretation from the executed computation, including an
    inactive local-gradient term, an algebraically scale-cancelling proxy, and
    the train--inference scaling mismatch;
    \item a completed six-condition, four-dataset, four-target, four-seed
    evaluation (384/384 runs), including native ERM and a fully replicated
    shared-learnable condition;
    \item source-only model selection and target-clustered uncertainty,
    with every named target and run seed reported separately; and
    \item a narrow negative conclusion: under this protocol, the full package
    shows no advantage over ERM, and separate source scales are macro-equivalent
    to one shared scale within the stated revision-stage margin.
\end{itemize}

\section{Related Work and Scope Boundary}
\label{sec:related}

\paragraph{Domain generalization and ERM.}
ERM minimizes average source risk \citep{vapnik1998statistical}. DomainBed
showed that controlled model selection and protocol matching can erase many
claimed DG gains \citep{gulrajani2021insearch}. Proposed DG objectives include
invariance, risk extrapolation, and distribution alignment
\citep{arjovsky2019invariant,krueger2021out,sun2016deep}. We therefore compare
against unmodified DomainBed ERM rather than importing literature numbers, and
we keep target-domain data out of the checkpoint score.

\paragraph{Genuine hyperbolic representations.}
Hyperbolic neural networks implement manifold-valued operations and
hyperbolic classifiers \citep{ganea2018hyperbolic}. MERU learns image--text
representations on a Lorentz hyperboloid \citep{desai2023meru}. Recent work
learns curvature within hyperbolic networks
\citep{fan2025curvaturelearning}, and DG-specific methods use hyperbolic state
space hallucination or cross-branch consistency
\citep{bi2025hyperbolic,disalvo2026hypcbc}. These are materially stronger
geometric parameterizations than the scalar rescaling audited here. They are
also different algorithms, tasks, architectures, and selection protocols, so
their published numbers are not valid matched controls for our experiment.
A controlled comparison would require implementing a manifold-valued encoder
or head in the same 384-cell protocol; the present evidence does not contain
such an experiment.

\begin{table}[t]
\centering
\small
\caption{The claim boundary. The last row is the only mechanism tested in this
paper. Hyperbolic in the other rows denotes manifold-valued operations or
representations, not merely a scalar applied to Euclidean features.}
\label{tab:scope-boundary}
\begin{tabular}{p{0.25\linewidth}p{0.34\linewidth}p{0.33\linewidth}}
\toprule
Family & Representation and prediction & Relation to this study \\
\midrule
Hyperbolic neural networks
\citep{ganea2018hyperbolic} &
Poincar\'e/M\"obius operations and a hyperbolic classifier &
Genuine manifold baseline; not implemented here \\
MERU and learned-curvature HNNs
\citep{desai2023meru,fan2025curvaturelearning} &
Lorentz or hyperbolic embeddings with explicit manifold geometry &
Stronger geometric parameterization; different protocol \\
Hyperbolic DG
\citep{bi2025hyperbolic,disalvo2026hypcbc} &
Hyperbolic representations embedded in DG-specific objectives &
Relevant recent direction; not a scalar-only ablation \\
\method{} (this audit) &
$h_\theta(x)/\sqrt{c_i}$ in training; Euclidean linear prediction on
$h_\theta(x)$ &
Curvature-inspired scalar feature scaling only \\
\bottomrule
\end{tabular}
\end{table}

\section{Method Under Test}
\label{sec:method}

\subsection{Executed feature scaling}

For source-loader slot $i$, the encoder produces
$h_i=h_\theta(x_i)\in\mathbb{R}^{2048}$. A free scalar $a_i$ is transformed to
\begin{equation}
c_i=\operatorname{softplus}(a_i)+10^{-4}, \qquad
z_i=\frac{h_i}{\sqrt{c_i}} .
\label{eq:scale}
\end{equation}
All $a_i$ and $a_g$ start at zero, hence
$c_i=c_g=\log2+10^{-4}=0.693247$ at initialization, giving a feature
multiplier $1/\sqrt{c}=1.201036$. In \textsc{MatchedControl} and
\textsc{Frozen} the scale optimizer's learning rate is set to zero, so $a_i$
and $a_g$ never move from zero and $c_i=c_g=0.693247$ is the constant value
used for every update of the entire run, not merely the initial value. In
\full{}, \novar{}, and \shared{}, $a_i$ (and, for \shared{}, the one shared
$a$) and $a_g$ are learned from this same starting point. A separate learned
$c_g$ defines $z_{g,i}=h_i/\sqrt{c_g}$. In a leave-one-domain-out run there
are three source loaders. Consequently, per-domain in the code means three
source-loader slots, whose semantic domain mapping changes with the held-out
target; there is no parameter for the unseen target.

The local supervised loss is
\begin{equation}
\mathcal L_{\mathrm{sup}}=\frac{1}{3}\sum_i
\operatorname{CE}(Wz_i+b,y_i).
\end{equation}
The code also forms domain-mean alignment
\begin{equation}
\mathcal L_{\mathrm{align}}=
\frac{1}{3}\sum_i\left\|\bar z_i-\sum_jw_j\bar z_j\right\|_2^2 ,
\end{equation}
where $\bar z_i=\frac{1}{|B_i|}\sum_{k\in B_i}z_{i,k}$ is the mean of the
scaled features over the $|B_i|=32$ examples in source minibatch $i$, and
$w_j=B_j/\sum_kB_k$ weights each source by its minibatch size (a
$10^{-8}$ denominator stabilizer is otherwise negligible). Every source batch
has 32 examples here, so $w_j=1/3$ for all $j$; the alignment term reduces to
pulling each domain mean toward the unweighted mean of the three domain means.
Pairwise scale dispersion is
\begin{equation}
\mathcal L_{\mathrm{var}}=\frac{1}{9}\sum_{i,j}(c_i-c_j)^2 ,
\end{equation}
and a global disagreement term
\begin{equation}
\mathcal L_{\mathrm{global}}=\frac{1}{3}\sum_i
\left|\operatorname{CE}(Wz_{g,i}+b,y_i)
-\operatorname{stopgrad}\!\left[
\operatorname{CE}(Wz_i+b,y_i)\right]\right|.
\end{equation}

\subsection{Intended versus executed auxiliary terms}
\label{sec:executed}

The executed code matters for interpretation. First, the quantity named the
norm/sharpness proxy is
\begin{equation}
\mathcal L_{\mathrm{norm}}=
\frac{1}{3}\sum_i c_i\,\operatorname{Var}\!\left(
\left\|h_i/\sqrt{c_i}\right\|_2\right)
=\frac{1}{3}\sum_i\operatorname{Var}(\|h_i\|_2),
\label{eq:cancellation}
\end{equation}
away from the inactive numerical clamp, where $\operatorname{Var}(\cdot)$ is
the population variance (divisor $|B_i|=32$, PyTorch \texttt{unbiased=False})
of the 32 per-example scaled-feature norms $\|h_{i,k}/\sqrt{c_i}\|_2$,
$k\in B_i$, computed separately within each source minibatch $i$ before the
three domain values are averaged. Because $\sqrt{c_i}$ is constant across the
examples in minibatch $i$, it factors out of the variance and squares back in
against the multiplying $c_i$, which is why the expression collapses to
$\operatorname{Var}(\|h_i\|_2)$: the term regularizes raw feature-norm
dispersion but its value does not depend on $c_i$, so it is not
curvature-sensitive and it is not SAM \citep{foret2021sharpness}.

Second, the intended local term differentiates each supervised loss with
respect to $a_i$. The implementation requests that gradient from a newly
indexed view that is not the view used to construct the forward scale.
Automatic differentiation therefore returns \texttt{None}, which is converted
to zero:
\begin{equation}
\mathcal L_{\mathrm{local}}^{\mathrm{executed}}=0.
\end{equation}
Every primary-run log records this term as zero. This is an implementation
defect, not evidence that a correctly implemented curvature sensitivity is
zero.

The executed total loss is
\begin{equation}
\mathcal L=
\mathcal L_{\mathrm{sup}}
+\alpha\mathcal L_{\mathrm{var}}
+\beta\mathcal L_{\mathrm{align}}
+\gamma\mathcal L_{\mathrm{norm}}
+\delta_\ell\mathcal L_{\mathrm{local}}
+\delta_g\mathcal L_{\mathrm{global}} .
\label{eq:objective}
\end{equation}
The full coefficients are
$(\alpha,\beta,\gamma,\delta_\ell,\delta_g)
=(0.2,0.5,0.05,0.1,0.1)$, but the local contribution is identically zero.
Finally, \texttt{predict()} applies $Wh_\theta(x)+b$, not
$Wz+b$. Thus scales act only through training; they are not used for target
prediction. With a linear head, a common global scale is also partly
absorbable into $W$.

\subsection{Six controlled conditions}
\label{sec:conditions}

The six conditions in Table~\ref{tab:conditions} play two different
inferential roles, and only two contrasts among them are primary claims of
this paper. \full{}$-$\erm{} and \novar{}$-$\shared{} are the two primary
suite contrasts defined in Section~\ref{sec:experiments} and carried through
Holm correction in Section~\ref{sec:results}. \textsc{MatchedControl} and
\textsc{Frozen} are secondary, descriptive diagnostics: no hypothesis test in
this paper treats a contrast involving either condition as a primary claim.
\textsc{MatchedControl} runs the \method{} update route with a constant,
non-learned feature rescaling and every auxiliary coefficient set to zero, so
it shows what the route alone costs or gains relative to \erm{} before any
\method{}-specific loss term is active. \textsc{Frozen} is the fixed-scale
counterpart of \novar{}: it keeps the same non-dispersion auxiliary losses
active but never lets the scales move, so comparing it with \novar{}
descriptively probes learned against fixed scaling. \textsc{Frozen} is not a
substitute for \shared{}, because it changes learnability rather than
per-source specificity; \novar{}$-$\shared{} is the condition pair that isolates
the domain-specificity question.
\begin{table}[!h]
\centering
\small
\caption{Completed conditions. All five non-ERM conditions use the same
\method{} update route. Only \erm{}, \full{}, \novar{}, and \shared{} enter the
paper's two primary contrasts; \textsc{MatchedControl} and \textsc{Frozen} are
descriptive diagnostics (see text).}
\label{tab:conditions}
\begin{tabular}{p{0.20\linewidth}p{0.73\linewidth}}
\toprule
Condition & Executed treatment \\
\midrule
\erm{} & Native DomainBed ERM; no scales or auxiliary losses. \\
\textsc{MatchedControl} & \method{} route; three fixed local scales and one fixed
global scale; all five coefficients zero. \\
\full{} & Three learned local scales and one learned global scale; full
coefficients in Equation~\ref{eq:objective}. \\
\textsc{Frozen} & All scales fixed; retains alignment, norm, local (inactive),
and global terms; $\alpha=0$. \\
\novar{} & Three learned local scales; same remaining objective as \full{} but
$\alpha=0$. \\
\shared{} & One learned local scale shared by all three source loaders plus the
separate learned global scale. Scale dispersion is structurally zero. \\
\bottomrule
\end{tabular}
\end{table}
\FloatBarrier

\section{Experimental Design}
\label{sec:experiments}

\paragraph{Benchmarks and why they are informative.}
We use PACS \citep{li2017deeper}, VLCS \citep{fang2013unbiased}, OfficeHome
\citep{venkateswara2017deep}, and TerraIncognita
\citep{beery2018recognition}. Their four environments emphasize different
shift types: depiction/style (PACS), acquisition and dataset source (VLCS),
depiction plus object context (OfficeHome), and camera location/background
(TerraIncognita). This diversity is useful for asking whether one scalar
specializes consistently across source environments. None of these datasets
provides ground-truth manifold curvature, hierarchy, or a known geometric
scale, so they cannot by themselves reveal that a representation's true
curvature differs by domain.

\paragraph{Protocol.}
Each environment is held out once; the other three are sources. Within every
environment, 80/20 splits are deterministic functions of the run seed and
environment index.
Seeds 0--3 vary data splits, initialization, minibatch order, and augmentation.
All methods use \texttt{timm} \texttt{resnet50.ram\_in1k} ImageNet weights, a
2,048-dimensional feature, a biased linear classifier, source batch size 32,
Adam with network learning rate $5\times10^{-5}$, zero weight decay, data
augmentation, 5,000 updates, and evaluation every 200 updates plus update
4999. BatchNorm is trainable. Learned scales have a separate Adam optimizer at
$10^{-4}$ and the \method{} route clips network and scale gradients to norm 5.
By \emph{native DomainBed ERM} we mean the unmodified upstream DomainBed
\texttt{ERM} class \citep{gulrajani2021insearch}, instantiated with the same
data, splits, augmentation, timm ResNet-50 backbone, and source-only selector
as every other condition: it concatenates the three source minibatches into
one batch, computes a single cross-entropy loss, and takes one Adam step with
one optimizer and no gradient clipping. It is the project's training and
evaluation harness around an otherwise untouched DomainBed ERM update rule,
not a claim that the surrounding software stack is an unmodified upstream
installation.

This last distinction prevents a curvature-only causal reading of
\full{}$-$\erm{}. Per update, \method{} calls the feature extractor six times:
three source-minibatch forward passes build the training objective, and,
after the optimizer step, three additional forward passes under
\texttt{torch.no\_grad()} recompute training accuracy for logging.
\texttt{no\_grad()} only suppresses gradient tracking; it does not put the
model in evaluation mode. The model is never switched to \texttt{eval()}
during \method{}'s update, so all six forward passes run in ordinary
training-mode BatchNorm, updating the running mean, running variance, and
batch counter exactly as any other training-mode forward pass. Native ERM
instead concatenates the three source minibatches and calls the feature
extractor once on all 96 examples, so its BatchNorm statistics update from one
training-mode forward pass per step rather than \method{}'s six. The
comparison therefore estimates the utility of the complete
implementation, including this BatchNorm-update asymmetry, not an isolated
curvature effect. The \novar{}$-$\shared{} comparison uses the same route on
both sides, so this asymmetry cancels out of that contrast.

\paragraph{Source-only checkpoint selection.}
For dataset $j$, held-out target $d$, seed $s$, and checkpoint $t$, selection
maximizes
\begin{equation}
q_{jds}(t)=\frac{1}{3}\sum_{r\neq d}
\operatorname{Acc}^{\mathrm{out}}_{jdsr}(t).
\label{eq:selection}
\end{equation}
Only source environments enter $q$; exact ties choose the earliest checkpoint.
Held-out-target accuracy is read after $t^*$ is fixed. The trainer did log
target curves, so the precise claim is that target accuracy is not used by the
frozen selection rule. Source \texttt{out} subsets inherit stochastic source
augmentation and are evaluated once per checkpoint, a limitation shared by
all conditions. Update-4999 results are a prespecified revision-stage
sensitivity endpoint.

\paragraph{Estimand.}
The scientific claim is about the equal-weighted macro over exactly these four
benchmarks (PACS, VLCS, OfficeHome, TerraIncognita) under this protocol, not
about a population or superpopulation of all DG tasks that these four
datasets are sampled from. We chose the suite for its heterogeneity of shift
type (Section~\ref{sec:experiments}), not as a random draw from a broader task
distribution, so every suite-level number below should be read as a
fixed-suite macro, not a generalization to unseen benchmarks.

\paragraph{Estimands and uncertainty.}
Let $Y_{mjds}$ be source-selected target accuracy (percentage points) for
method $m$. For contrast $A-B$,
\begin{equation}
g_{jds}=Y_{Ajds}-Y_{Bjds},\quad
\bar g_{jd}=\frac{1}{4}\sum_s g_{jds},\quad
\Delta_j=\frac{1}{4}\sum_d\bar g_{jd}.
\end{equation}
Seeds are averaged within target first, and targets are averaged within
dataset second; the suite-level procedures below use only the four resulting
dataset means $\Delta_j$ and do not separately repropagate seed- or
target-level variance into the suite interval. That within-dataset
variability is real and is reported descriptively (per-target effects,
between-target SD, and within-target seed SD all appear in the supplement),
but it enters the suite-level test only through the four dataset means it
produced, not as an additional variance component.

The four named target effects $\bar g_{jd}$ are the per-dataset inferential
units. We report an exact two-sided sign-flip test over all $2^4=16$ target
assignments, a $t_3$ interval, and a four-cluster bootstrap sensitivity. The
sign-flip test is exact conditional on treating the four target-level sign
flips as exchangeable under the joint no-effect null (every one of the 16
sign patterns equally likely if $\bar g_{jd}$ has a distribution symmetric
about zero under the null); it is a working symmetry assumption for a
four-cluster benchmark suite, not a claim that the datasets were themselves
randomly assigned to conditions, and it does not test whether dataset-level
effects cancel under a suite-macro-only null. The smallest possible two-sided
exact $p$ is .125 with four clusters. At suite level,
$\Delta_{\mathrm{macro}}=\frac14\sum_j\Delta_j$ and the four dataset effects
are the clusters. The two primary suite contrasts, \full{}$-$\erm{} and
\novar{}$-$\shared{}, form one Holm family \citep{holm1979simple}. Per-dataset
tests are an eight-test sensitivity family.

We declare the exact sign-flip test with Holm correction as the primary
inferential procedure for both suite-level difference claims; the $t_3$
interval and the bootstrap sensitivity are secondary, working-model summaries
that assume approximate normality or resampling validity with only four
clusters and do not override the primary test. Where the two disagree, as for
\full{}$-$\erm{} (Section~\ref{sec:results}), the Holm-adjusted exact result
governs the paper's claim.

Practical equivalence uses TOST \citep{schuirmann1987comparison} with a
revision-stage smallest effect of practical interest (SESOI) of $\pm1\pp$,
chosen after the submitted results were known but before the added ERM/shared
grid completed. Margins of $\pm0.5$ and $\pm2\pp$ are sensitivities. TOST,
like the $t_3$ interval, is a secondary working-model summary conditional on
the same four-cluster suite macro; we do not interpret a nonsignificant
difference-test result as equivalence, and equivalence conclusions are
likewise scoped to this fixed four-dataset macro.

\section{Results}
\label{sec:results}

\subsection{Completed six-condition study}

All 384 planned runs completed.
Table~\ref{tab:accuracy} reports equal-target macro accuracies. Every named
target and seed, along with between-target and within-target run-seed
variability, appears in the supplement.

\begin{table}[t]
\centering
\small
\caption{Source-validation-selected target accuracy (\%). Each cell averages
four named targets after averaging four run seeds within target. Bold marks
the highest point estimate in a dataset, not a significance claim.}
\label{tab:accuracy}
\begin{tabular}{lrrrr}
\toprule
Condition & PACS & VLCS & OfficeHome & TerraInc. \\
\midrule
\erm{}                 & 85.88 & \textbf{77.19} & \textbf{70.21} & \textbf{47.13} \\
\textsc{MatchedControl} & 84.22 & 75.38 & 70.05 & 44.14 \\
\full{}                & 84.87 & 76.42 & 68.52 & 44.17 \\
\textsc{Frozen}      & \textbf{86.14} & 75.84 & 67.81 & 44.20 \\
\novar{}               & 85.28 & 76.17 & 68.36 & 43.70 \\
\shared{}              & 85.55 & 76.56 & 68.17 & 43.75 \\
\bottomrule
\end{tabular}
\end{table}

\subsection{Primary utility and domain-specificity contrasts}

\begin{table}[t]
\centering
\small
\caption{Primary suite-macro contrasts. Exact $p$-values sign-flip the four
dataset effects; Holm adjusts the two primary tests. TOST uses the
$\pm1\pp$ revision-stage SESOI.}
\label{tab:primary}
\setlength{\tabcolsep}{3pt}
\begin{tabular}{p{0.15\linewidth}p{0.22\linewidth}p{0.15\linewidth}p{0.10\linewidth}p{0.25\linewidth}}
\toprule
Contrast & Effect [95\% CI] & Exact / Holm $p$ & TOST $p$ & Interpretation \\
\midrule
Full $-$ ERM &
$-1.61$ $[-3.17,-0.04]$ & .125 / .250 & .848 &
No adjusted evidence; equivalence not demonstrated \\
NoVar $-$ Shared &
$-0.13$ $[-0.53,0.27]$ & .375 / .375 & .003 &
$\pm1\pp$ source-selected macro-equivalent \\
\bottomrule
\end{tabular}
\end{table}

\paragraph{Whole-package utility.}
\full{}$-$\erm{} is $-1.01$, $-0.77$, $-1.69$, and $-2.96\pp$ on PACS, VLCS,
OfficeHome, and TerraIncognita. Thus all four point estimates favor ERM.
The suite $t_3$ interval excludes zero, but with four dataset clusters the
primary exact test is $p=.125$ and Holm-adjusted $p=.250$; per
Section~\ref{sec:experiments}, this primary Holm-adjusted result, not the
secondary $t_3$ interval, governs the paper's claim. We therefore do not
claim adjusted evidence of a difference. The compatible range also does not
support equivalence of the packages. Relative to the \method{}-path matched control,
\full{} changes the four dataset means by $+0.65$, $+1.04$, $-1.53$, and
$+0.03\pp$; this secondary pattern is mixed.

\paragraph{Are separate source scales necessary?}
The clean \novar{}$-$\shared{} effects are $-0.27$, $-0.39$, $+0.19$, and
$-0.05\pp$ across the same datasets. The suite effect is $-0.13\pp$ with
90\% CI $[-0.43,0.17]$, wholly inside $\pm1\pp$ (TOST $p=.003$) and also
inside the $\pm0.5\pp$ sensitivity margin (TOST $p=.031$). This supports a
specific conclusion: for the source-selected, equal-dataset/equal-target
macro estimand, three source-loader scales offer no practically meaningful
advantage over one shared learned scale. We do not extend equivalence to any
individual dataset or target, whose intervals are wider.

\begin{figure}[t]
\centering
\begin{minipage}[t]{0.49\linewidth}
\centering
\includegraphics[width=\linewidth]{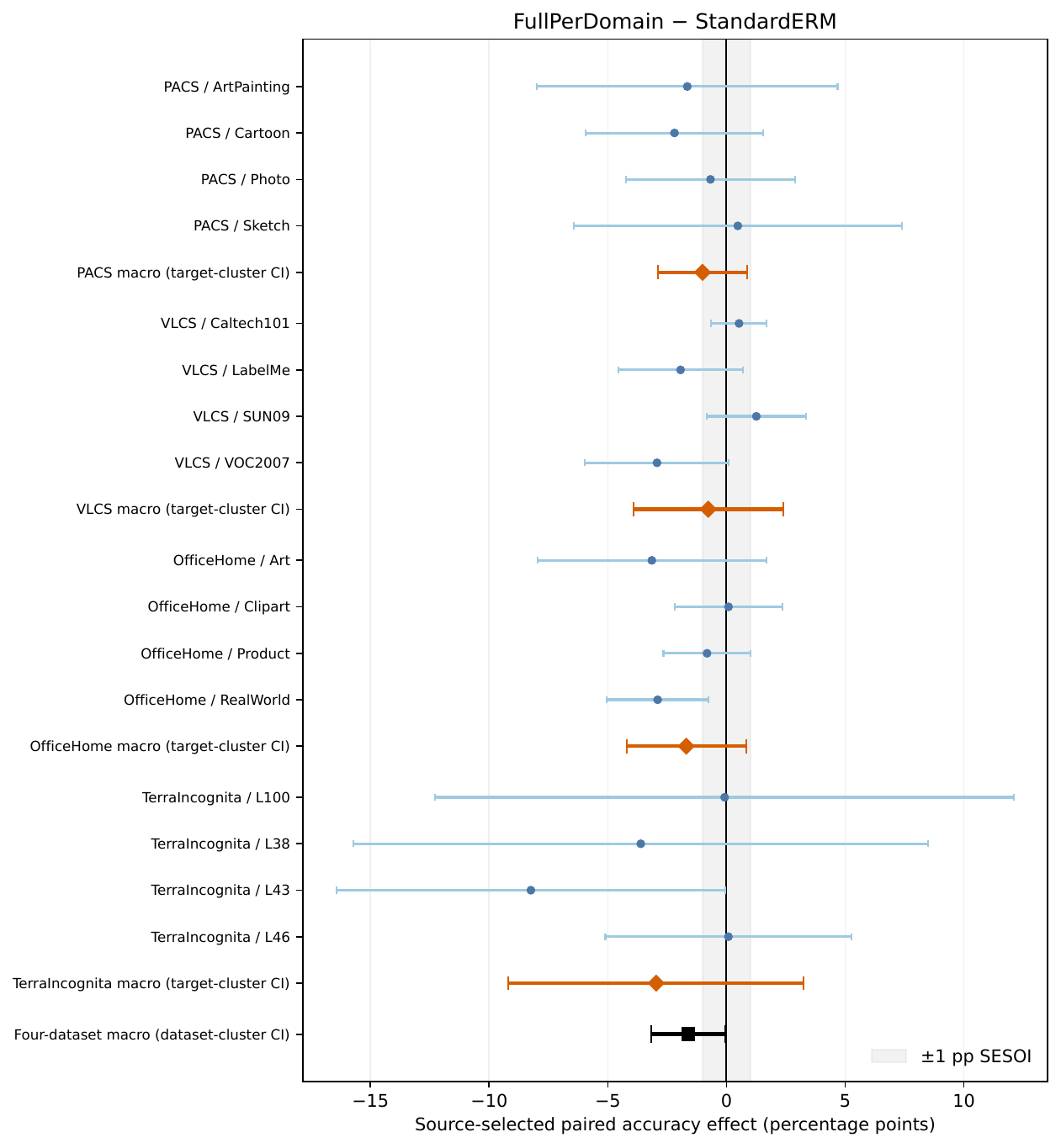}
\end{minipage}\hfill
\begin{minipage}[t]{0.49\linewidth}
\centering
\includegraphics[width=\linewidth]{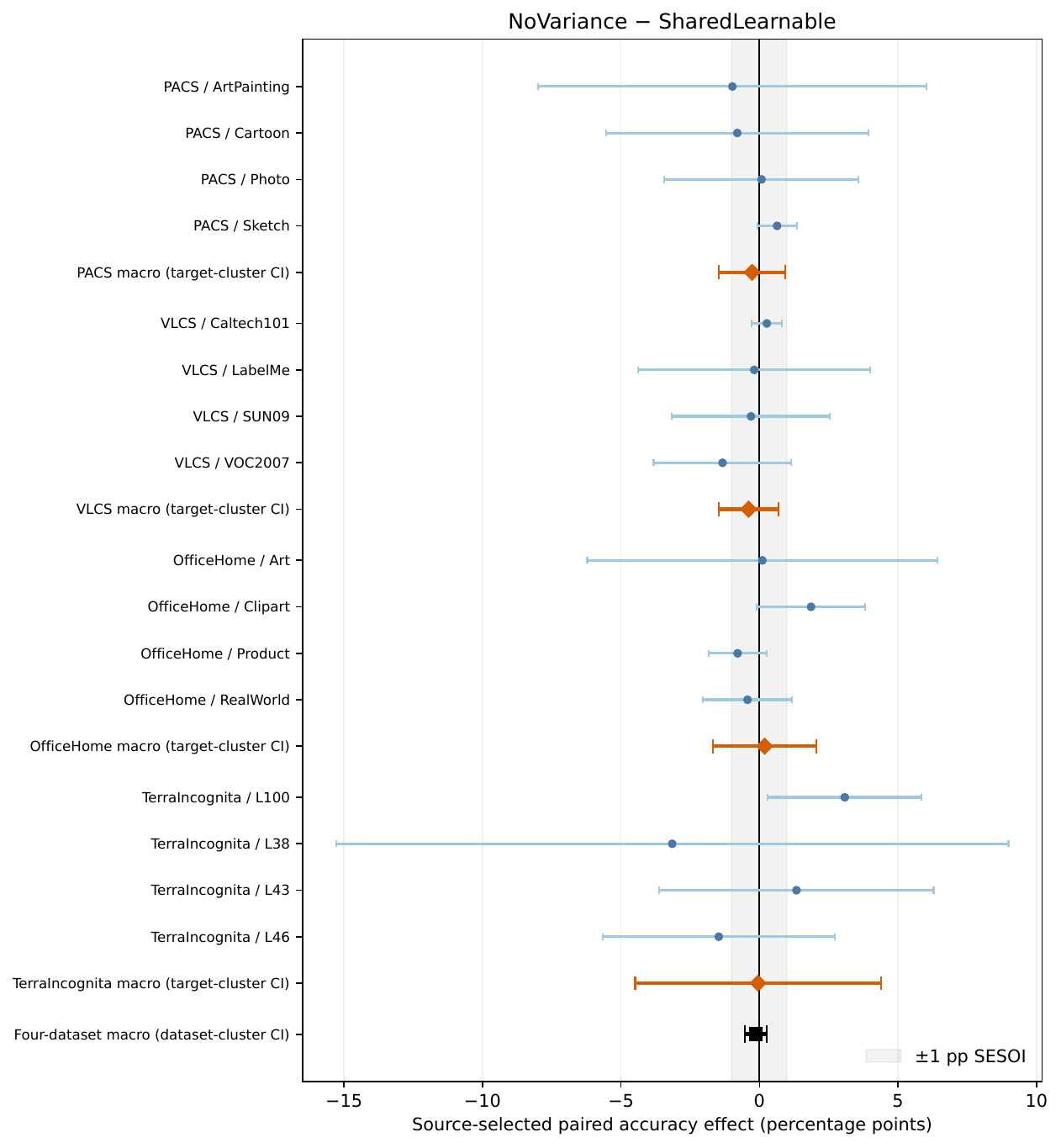}
\end{minipage}
\caption{Paired effects for the two primary contrasts. Blue intervals are
$t_3$ intervals across four paired run seeds within a named target; orange
intervals cluster four target means within a dataset; black intervals cluster
the four fixed dataset means. The visible target heterogeneity is why
seed-by-target cells are not treated as 16 independent domain draws. These
intervals describe this fixed, heterogeneous benchmark suite, not all DG
tasks. Exact values and run-seed SDs are in the supplement.}
\label{fig:forests}
\end{figure}

\paragraph{Checkpoint sensitivity.}
Source selection improves over update 4999 in 20 of 24
condition-by-dataset cells, by $0.97\pp$ on average. At update 4999,
\novar{}$-$\shared{} is $-0.32\pp$ with 90\% CI $[-1.13,0.50]$ and TOST
$p=.072$; equivalence is not established. We therefore restrict the
equivalence conclusion to the declared source-selected endpoint.

\paragraph{Mechanistic audit.}
The local-gradient regularizer is zero in all primary logs, and
Equation~\ref{eq:cancellation} shows that the norm proxy cannot identify
curvature. The close performance of \novar{} and \shared{} is consequently
consistent with a weakly identified, partly absorbable training-time scale.
It should not be interpreted as evidence that a correctly implemented
manifold-valued curvature parameter would behave similarly.

\section{Discussion}
\label{sec:discussion}

The empirical answer is narrower than the original geometric motivation.
First, the complete implementation provides no evidence of improvement over
native ERM under the tested protocol: its four dataset effects are negative,
while exact
small-cluster inference prevents a strong adjusted difference claim. Second,
within the common \method{} route, separately learned source scales are
macro-equivalent to one shared learned scale at the stated source-selected
endpoint. This directly answers the parameterization question that Frozen
could not answer: Frozen changes learnability and is not a substitute for a
shared-learnable condition.

The dataset suite is useful for stress-testing scale specialization because it
spans artistic style, dataset acquisition, object context, and camera-location
shifts. The heterogeneity in Figure~\ref{fig:forests} also shows why target
identity dominates some uncertainty summaries. But benchmark diversity is not
geometric supervision. Without known hierarchies, distances, or target
curvatures, a learned scalar cannot be validated as physical or sectional
curvature merely because it differs between environments.

The comparison to recent hyperbolic methods is therefore conceptual rather
than numerical. A genuine follow-up should keep the present splits, seeds,
backbone, budget, and source-only selector while replacing the Euclidean head
with a correctly implemented Poincar\'e or Lorentz representation and
hyperbolic classifier. Fixed- versus learned-curvature versions would then
separate manifold geometry from scalar flexibility. That experiment is
scientifically worthwhile, but adding incomplete MERU runs or post-hoc
Poincar\'e projections would not answer it; we exclude those diagnostics from
the evidentiary claims.

\section{Limitations and Broader Impact}
\label{sec:limitations}

\paragraph{Limitations.}
There are only four named target clusters per benchmark and four dataset
clusters at suite level, so exact tests are coarse and per-target intervals can
be wide. The study uses one ImageNet-pretrained ResNet-50 recipe and four
classification benchmarks. The SESOI and source-selected endpoint were fixed
during revision rather than preregistered before the original study. Source
validation is stochastic, and target curves existed in logs even though they
do not enter the frozen selection rule. Native ERM and \method{} have different
BatchNorm/update routes. Most importantly, the implementation has an inactive
local term, a scale-cancelling proxy, and unscaled inference, and the study
does not include genuine manifold-valued training. These facts narrow the
claim; they do not invalidate the negative result for the executed package.

\paragraph{Broader impact.}
This work uses standard image benchmarks and introduces no new human-subject
data. Its main impact is methodological: reporting a carefully bounded
negative result can prevent a scalar Euclidean implementation from being
mistaken for evidence about hyperbolic learning as a whole. The same care is
needed in the other direction; our results should not be cited to dismiss
genuine hyperbolic representations.

\section{Conclusion}

We evaluated a specific curvature-inspired scalar feature-scaling package,
not general hyperbolic representation learning. Across 384 completed runs,
the full implementation showed no advantage over native ERM under the
declared protocol, and its four dataset point estimates were lower. Within the
shared update route, separate source-loader scales were equivalent to one
shared learned scale for the source-selected suite macro within a
revision-stage $\pm1\pp$ margin, but not for every dataset, target, or
final-step endpoint. Together with the implementation audit, these results
locate the negative finding precisely: it applies to weakly identified
training-time Euclidean rescaling with this ResNet-50 protocol and should not
be generalized to manifold-valued hyperbolic models.

\small
\bibliography{main}

@article{wang2022generalizing,
  title={Generalizing to Unseen Domains: A Survey on Domain Generalization},
  author={Wang, Jindong and Lan, Cuiling and Liu, Chang and Ouyang, Yidong and Qin, Tao and Lu, Wang and Chen, Yiqiang and Zeng, Wenjun and Yu, Philip S.},
  journal={IEEE Transactions on Knowledge and Data Engineering},
  volume={35},
  number={8},
  pages={8052--8072},
  year={2023},
  doi={10.1109/TKDE.2022.3178128}
}

@inproceedings{gulrajani2021insearch,
  title={In Search of Lost Domain Generalization},
  author={Gulrajani, Ishaan and Lopez-Paz, David},
  booktitle={International Conference on Learning Representations},
  year={2021}
}

@book{vapnik1998statistical,
  title={Statistical Learning Theory},
  author={Vapnik, Vladimir N.},
  publisher={Wiley},
  address={New York},
  year={1998}
}

@article{arjovsky2019invariant,
  title={Invariant Risk Minimization},
  author={Arjovsky, Martin and Bottou, L{\'e}on and Gulrajani, Ishaan and Lopez-Paz, David},
  journal={arXiv preprint arXiv:1907.02893},
  year={2019}
}

@inproceedings{krueger2021out,
  title={Out-of-Distribution Generalization via Risk Extrapolation},
  author={Krueger, David and Caballero, Ethan and Jacobsen, Joern-Henrik and Zhang, Amy and Binas, Jonathan and Zhang, Dinghuai and Le Priol, Remi and Courville, Aaron},
  booktitle={International Conference on Machine Learning},
  pages={5815--5826},
  year={2021},
  organization={PMLR}
}

@inproceedings{sun2016deep,
  title={Deep {CORAL}: Correlation Alignment for Deep Domain Adaptation},
  author={Sun, Baochen and Saenko, Kate},
  booktitle={European Conference on Computer Vision},
  pages={443--450},
  year={2016}
}

@inproceedings{ganea2018hyperbolic,
  title={Hyperbolic Neural Networks},
  author={Ganea, Octavian-Eugen and B{\'e}cigneul, Gary and Hofmann, Thomas},
  booktitle={Advances in Neural Information Processing Systems},
  volume={31},
  year={2018}
}

@article{fan2025curvaturelearning,
  title={Curvature Learning for Generalization of Hyperbolic Neural Networks},
  author={Fan, Xiaomeng and Wu, Yuwei and Gao, Zhi and Harandi, Mehrtash and Jia, Yunde},
  journal={International Journal of Computer Vision},
  volume={133},
  number={12},
  pages={8489--8525},
  year={2025},
  doi={10.1007/s11263-025-02567-4}
}

@inproceedings{bi2025hyperbolic,
  title={Learning Fine-Grained Domain Generalization via Hyperbolic State Space Hallucination},
  author={Bi, Qi and Yi, Jingjun and Zhan, Haolan and Ji, Wei and Xia, Gui-Song},
  booktitle={Proceedings of the AAAI Conference on Artificial Intelligence},
  volume={39},
  pages={1853--1861},
  year={2025},
  doi={10.1609/aaai.v39i2.32180}
}

@article{disalvo2026hypcbc,
  title={{HypCBC}: Domain-Invariant Hyperbolic Cross-Branch Consistency for Generalizable Medical Image Analysis},
  author={Di Salvo, Francesco and Doerrich, Sebastian and Alle, Jonas and Ledig, Christian},
  journal={Transactions on Machine Learning Research},
  year={2026},
  note={arXiv:2602.03264}
}

@inproceedings{desai2023meru,
  title={Hyperbolic Image-Text Representations},
  author={Desai, Karan and Nickel, Maximilian and Rajpurohit, Tanmay and Johnson, Justin and Vedantam, Ramakrishna},
  booktitle={International Conference on Machine Learning},
  pages={7694--7731},
  year={2023},
  organization={PMLR}
}

@inproceedings{foret2021sharpness,
  title={Sharpness-Aware Minimization for Efficiently Improving Generalization},
  author={Foret, Pierre and Kleiner, Ariel and Mobahi, Hossein and Neyshabur, Behnam},
  booktitle={International Conference on Learning Representations},
  year={2021}
}

@inproceedings{li2017deeper,
  title={Deeper, Broader and Artier Domain Generalization},
  author={Li, Da and Yang, Yongxin and Song, Yi-Zhe and Hospedales, Timothy M.},
  booktitle={Proceedings of the IEEE International Conference on Computer Vision},
  pages={5542--5550},
  year={2017}
}

@inproceedings{fang2013unbiased,
  title={Unbiased Metric Learning: On the Utilization of Multiple Datasets and Web Images for Softening Bias},
  author={Fang, Chen and Xu, Ye and Rockmore, Daniel N.},
  booktitle={Proceedings of the IEEE International Conference on Computer Vision},
  pages={1657--1664},
  year={2013}
}

@inproceedings{venkateswara2017deep,
  title={Deep Hashing Network for Unsupervised Domain Adaptation},
  author={Venkateswara, Hemanth and Eusebio, Jose and Chakraborty, Shayok and Panchanathan, Sethuraman},
  booktitle={Proceedings of the IEEE Conference on Computer Vision and Pattern Recognition},
  pages={5018--5027},
  year={2017}
}

@inproceedings{beery2018recognition,
  title={Recognition in Terra Incognita},
  author={Beery, Sara and Van Horn, Grant and Perona, Pietro},
  booktitle={Proceedings of the European Conference on Computer Vision},
  pages={456--473},
  year={2018}
}

@article{holm1979simple,
  title={A Simple Sequentially Rejective Multiple Test Procedure},
  author={Holm, Sture},
  journal={Scandinavian Journal of Statistics},
  volume={6},
  number={2},
  pages={65--70},
  year={1979}
}

@article{schuirmann1987comparison,
  title={A Comparison of the Two One-Sided Tests Procedure and the Power Approach for Assessing the Equivalence of Average Bioavailability},
  author={Schuirmann, Donald J.},
  journal={Journal of Pharmacokinetics and Biopharmaceutics},
  volume={15},
  number={6},
  pages={657--680},
  year={1987},
  doi={10.1007/BF01068419}
}
\bibliographystyle{tmlr}

\end{document}